\documentclass[conference]{IEEEtran}
\IEEEoverridecommandlockouts
\usepackage{hyperref} 
\usepackage{cite}
\usepackage{amsmath,amssymb,amsfonts}
\usepackage{algorithmic}
\usepackage{graphicx}
\usepackage{textcomp}
\usepackage{xcolor}
\usepackage{eso-pic} % added this package for copyright
\def\BibTeX{{\rm B\kern-.05em{\sc i\kern-.025em b}\kern-.08em
    T\kern-.1667em\lower.7ex\hbox{E}\kern-.125emX}}
\begin{document}

\begin{titlepage}
    \centering
    \vspace*{4cm}
    % {\large This work has been submitted to the IEEE for possible publication.\\
    % Copyright may be transferred without notice, after which this version may no longer be accessible.}
   
    {\large © 2025 IEEE.  Personal use of this material is permitted.  \\ 
    Permission from IEEE must be obtained for all other uses, in any current or future media, including reprinting/republishing this material for advertising or promotional purposes, creating new collective works, for resale or redistribution to servers or lists, or reuse of any copyrighted component of this work in other works.}

    \vspace{2em}

    {\large 
    Preprint version.\\The final version is published in the Proceedings of CompSysTech’25, IEEE Xplore, 2025.}
    \vspace{2em}
    {\large \\Related DOI: \href{https://doi.org/10.1109/CompSysTech65493.2025.11137220}{https://doi.org/10.1109/CompSysTech65493.2025.11137220}}

    \vfill
\end{titlepage}

\title{Classification of Tents in Street Bazaars Using CNN}
% \thanks{} 

\author{\IEEEauthorblockN{1\textsuperscript{st} Azamat Ibragimov}
\IEEEauthorblockA{\textit{dept. of Applied Mathematics and Informatics} \\
\textit{Ala-Too International university}\\
Bishkek, Kyrgyzstan \\
0009-0007-8008-015X} 
\and

\hspace*{-0.8em}\IEEEauthorblockN{2\textsuperscript{nd} Ruslan Isaev}
\IEEEauthorblockA{\hspace*{-0.8em}\textit{dept. of Computer Science} \\
\hspace*{-0.8em}\textit{Ala-Too International university}\\
\hspace*{-0.8em}Bishkek, Kyrgyzstan \\
\hspace*{-0.8em}\textit{dept. of Computer Science} \\
\hspace*{-0.8em}\textit{Paragon International University} \\
\hspace*{-0.8em}Penh, Cambodia \\
\hspace*{-0.8em}0000-0003-4426-8837}
\and

\IEEEauthorblockN{3\textsuperscript{rd} Remudin Reshid Mekuria}
\IEEEauthorblockA{\textit{dept. of Applied Mathematics and Informatics} \\
\textit{Ala-Too International university}\\
Bishkek, Kyrgyzstan \\
0000-0001-6207-8143}
\and
\hspace*{6.5em}\IEEEauthorblockN{4\textsuperscript{th} Gulnaz Gimaletdinova}
\IEEEauthorblockA{\hspace*{6.5em}\textit{dept. of Applied Mathematics and Informatics} \\
\hspace*{6.5em}\textit{Ala-Too International university}\\
\hspace*{6.5em}Bishkek, Kyrgyzstan \\
\hspace*{6.5em}0009-0001-9104-0918}
\and

\IEEEauthorblockN{5\textsuperscript{th} Dim Shaiakhmetov}
\IEEEauthorblockA{\textit{dept. of Computer Science} \\
\textit{Ala-Too International university}\\
Bishkek, Kyrgyzstan \\
0009-0006-3534-960X}
}

% \AddToShipoutPictureFG*{%
%   \AtPageLowerLeft{%
%     \put(\LenToUnit{0.7in},\LenToUnit{0.5in}){%
%       \footnotesize 979-8-3315-4322-8/25/\$31.00 \copyright2025 IEEE%
%     }%
%   }%
% }

\maketitle

\begin{abstract}
This research paper proposes an improved deep learning model for classifying tents in street bazaars, comparing a custom Convolutional Neural Network (CNN) with EfficientNetB0. This is a critical task for market organization with a tent classification, but manual methods in the past have been inefficient. Street bazaars represent a vital economic hub in many regions, yet their unstructured nature poses significant challenges for the automated classification of market infrastructure, such as tents. In Kyrgyzstan, more than a quarter of the country's GDP is derived from bazaars. While CNNs have been widely applied to object recognition, their application to bazaar-specific tasks remains underexplored. Here, we build upon our original approach by training on an extended set of 126 original photographs that were augmented to generate additional images. This dataset is publicly available for download on Kaggle. A variety of performance metrics, such as accuracy, precision, recall, F1 score, and mean average precision (mAP), were used to assess the models comparatively, providing a more extensive analysis of classification performance.

The results show that the CNN custom model achieved 92.8\% accuracy, and EfficientNetB0 showed 98.4\% accuracy results, confirming the effectiveness of transfer learning in the bazaar image classification. Also, when analyzing the confusion matrix, the analysis reveals the weaknesses and strengths of each model. These findings suggest that using a pre-trained model such as EfficientNetB0 significantly improves classification accuracy and generalization.
\end{abstract}


\section{Introduction}
Bazaars are a key component of the informal economy in Central Asia, with estimates suggesting that the informal economy constitutes a significant portion of GDP, ranging from 26\% in Kyrgyzstan to 35\% in Armenia \cite{Karrar2019}. Bazaars facilitate regional trade and cross-border cooperation, enhancing economic opportunities for local populations. They serve as critical nodes in trade networks connecting Central Asia with global markets, including China, India, and Turkey \cite{Kaminski2012}.

In most regions, especially in Central Asia, street markets are an important element of the economic system. Here, sellers use special structures such as tents to trade various goods, including groceries, dried fruits, and household items. The classification of such tents is of great importance for the organization of retail space, analysis of the structure of trade and automation of inventory management. However, classical classification methods that are performed manually are not always effective, time-consuming, and are prone to human error.
With the development of technology and the availability of computer vision methods, deep learning models have become a promising solution for automating the process of classifying
objects related to the market.

In our previous research \cite{Ibragimov2024}, we developed a custom CNN model, and although the model has shown promising results, its generalization and reliability can still be improved. In this study, we expand our approach by implementing EfficientNetB0, an advanced deep learning architecture optimized for both efficiency and accuracy. By using transfer learning, we aim to improve classification efficiency.

\section{Literature review}
Deep learning, particularly CNNs, has shown strong potential for object detection and classification in complex environments like street bazaars. This section reviews CNN-based classification, transfer learning for small datasets, and data augmentation techniques to improve model performance.

CNN-based object Detection and Classification are widely used for object classification due to their ability to extract hierarchical features from images. Traditional CNN models, such as VGG and ResNet, have been employed in various retail and market classification tasks, but their performance declines in environments with high occlusion and visual clutter. Recent studies, including Muñoz et al. \cite{Munoz2024}, highlight the effectiveness of YOLO-based architectures for real-time object detection in retail environments. However, in bazaars, the presence of overlapping tents and inconsistent tent structures makes classification more challenging. Srivastava \cite{Srivastava} introduces fine-grained classification methods that leverage local feature extraction, which could be beneficial in distinguishing different types of tents based on structural features.

\subsection{Transfer Learning for Small Datasets}
A major limitation in bazaar classification tasks is the scarcity of labeled datasets. Transfer learning offers a solution by utilizing pre-trained models on large datasets and fine-tuning them for specific tasks. EfficientNetB0, as used in this study, is an example of a lightweight model that benefits from transfer learning, achieving high classification accuracy with minimal data. Wei et al. \cite{Wei2020} emphasize that models pre-trained on diverse datasets, such as ImageNet, can significantly improve accuracy in specialized applications. In particular, Srivastava \cite{Srivastava} demonstrates that models trained on social media datasets, like Instagram-pretrained ResNext-WSL, exhibit greater robustness to real-world distortions, an advantage when working with street bazaar images taken under non-uniform conditions.

\subsection{Data Augmentation for Model Improvement}
Due to the limited availability of training data for tents in street bazaars, data augmentation plays a critical role in enhancing model generalization. Techniques such as rotation, scaling, color jittering, and synthetic image generation help improve model robustness to variations in tent appearances. Varatharasan et al. \cite{Varatharasan2025} propose a novel approach using DeepLab-based segmentation and Chroma-keying to create training samples with varying backgrounds, which aligns with the need to simulate real-world bazaar conditions. Wei et al. \cite{Wei2020} further explore the application of GANs for generating synthetic datasets, reducing the dependency on extensive manual labeling. In this study, the dataset used has been publicly released on Kaggle, allowing for further improvements and benchmarking by the research community. As an alternative, an image-to-image technique can be used for data augmentation to generate a lot of images that can be used for training \cite{Toktosunova2024}.

It is important to show possibilities in training and implementation of neural network models in authentic, cultural and business fields such as markets and street bazaars, which have a high impact on GPD and feeds the people in an evolving region. Digitalizing bazaars through AI, GIS, and augmented technologies can raise the cultural phenomena to a qualitatively new level.

\section{Methodology}

This study presents an approach to tent classification in street bazaars using deep learning-based image classification. We compare the performance of two architectures: a custom CNN trained from initialization and EfficientNetB0, a pre-trained model optimized for efficiency and accuracy. This section details the preparation of the data set, model architectures, training procedures, and evaluation methods used in our experiments.

\subsection{Dataset Preparation}
\begin{figure}[h]
    \centering
    \includegraphics[width=0.9\linewidth]{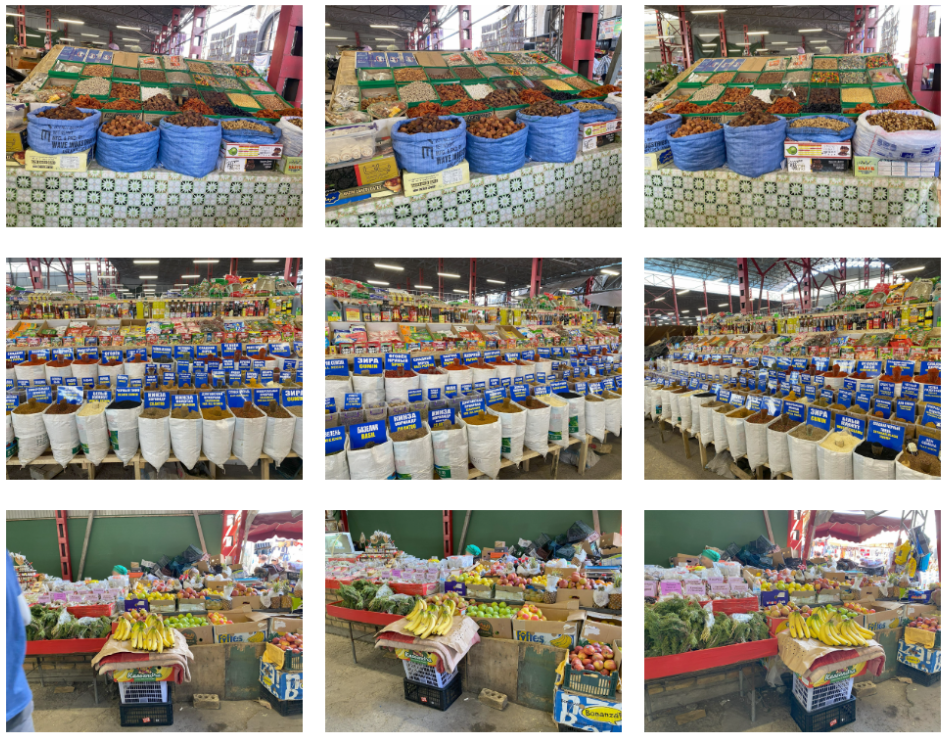}
    \caption{Example of photos}
    \label{fig:orig_photos}
\end{figure}

\begin{figure}[h]
    \centering
    \includegraphics[width=0.9\linewidth]{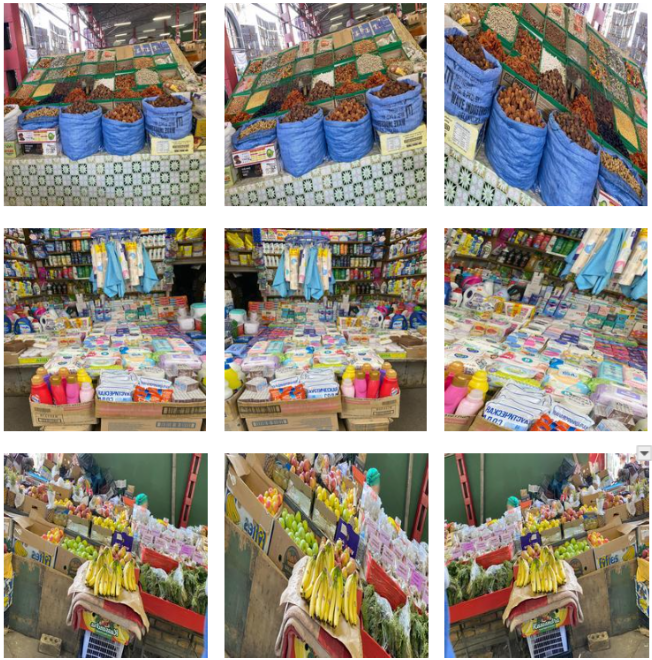}
    \caption{Example of augmented photos}
    \label{fig:aug_photos}
\end{figure}

The dataset used in this study contains 126 original images of tents divided into 6 classes: dried fruits, foodstuff, fruit and vegetables, household goods, meat, spices. Captured from various angles in street bazaar, you can see the examples in Figure~\ref{fig:orig_photos}. Due to the limited number of raw images, data augmentation was applied to artificially increase the dataset size and improve model generalization. As a result, each class now contains 421 images, leading to a total of 2,526 augmented images. Examples are shown in Figure \ref{fig:aug_photos}. Augmentation techniques included rotating images by ±25°, horizontal flipping, adjusting brightness for varying lighting, and zooming and cropping to introduce scale variations. 


The dataset was split into \textbf{80\% training} and \textbf{20\% validation} subsets, ensuring a balanced representation of all categories. 

\subsection{Model Architectures}
To evaluate the effectiveness of deep learning in tent classification, we implemented and compared two architectures:

\subsubsection{Custom CNN}
\begin{figure}[h]
    \centering
    \includegraphics[width=0.9\linewidth]{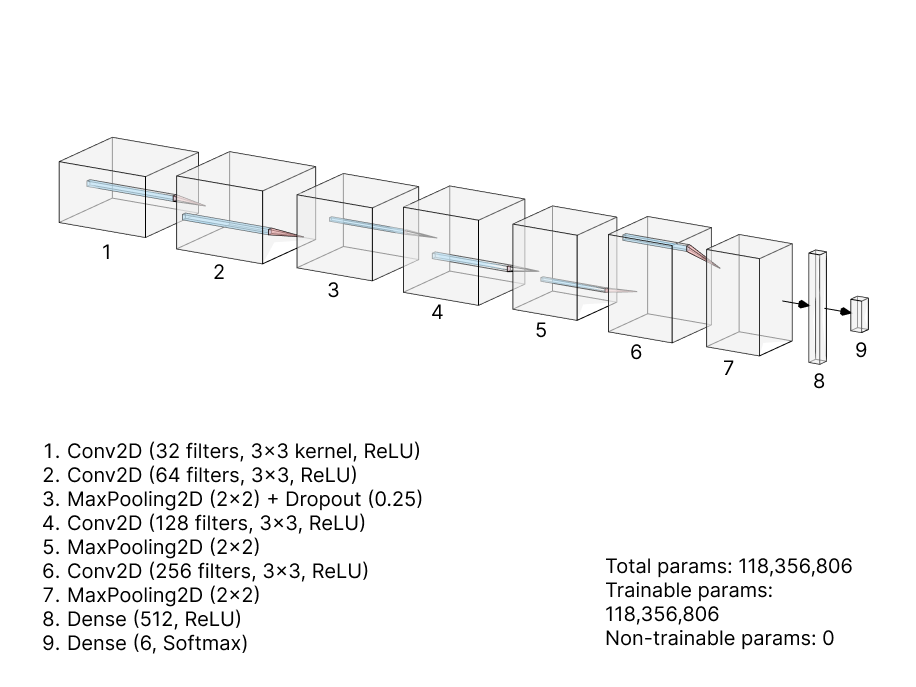}
    \caption{Visualization of Custom CNN model}
    \label{fig:model}
\end{figure}

A CNN was designed with multiple convolutional and pooling layers, followed by fully connected layers. You can see the visualization of the model in the Figure~\ref{fig:model}. 

This model was trained from initialization, requiring a substantial number of training epochs and careful regularization to prevent overfitting.

\subsubsection{EfficientNetB0 (Transfer Learning)}
We used EfficientNetB0, a pre-trained CNN model on ImageNet, for feature extraction. The main changes included removing the original classification head, adding a Global Average Pooling (GAP) layer, incorporating fully connected layers with ReLU and Dropout, and using a softmax output layer to classify images into six tent categories.

EfficientNetB0 was first trained with frozen base layers, optimizing only the newly added classification layers. After 10 epochs, the entire network was fine-tuned to improve accuracy.

\subsection{Training Procedure and Hyperparameter Optimization}
Both models were trained using TensorFlow and Keras with a batch size of 64 and 32, respectively, and 20 epochs. The custom CNN used categorical crossentropy as the loss function, Adam optimizer with a learning rate of $1 \times 10^{-4}$, and 50\% dropout for regularization. EfficientNetB0 also used categorical crossentropy, Adam optimizer with the same learning rate, and applied L2 regularization ($\lambda = 0.01$) along with 50\% dropout. Training for EfficientNetB0 was done in two phases. In Phase 1, the base layers were frozen, and only the classifier was trained for 10 epochs. In Phase 2, all layers were unfrozen and fine-tuned for 20 more epochs.

The models were evaluated on a separate test set containing only original images, ensuring that performance was assessed under realistic conditions.

\section{Results}

\begin{figure*}[t]
    \centering
    \includegraphics[height=7cm]{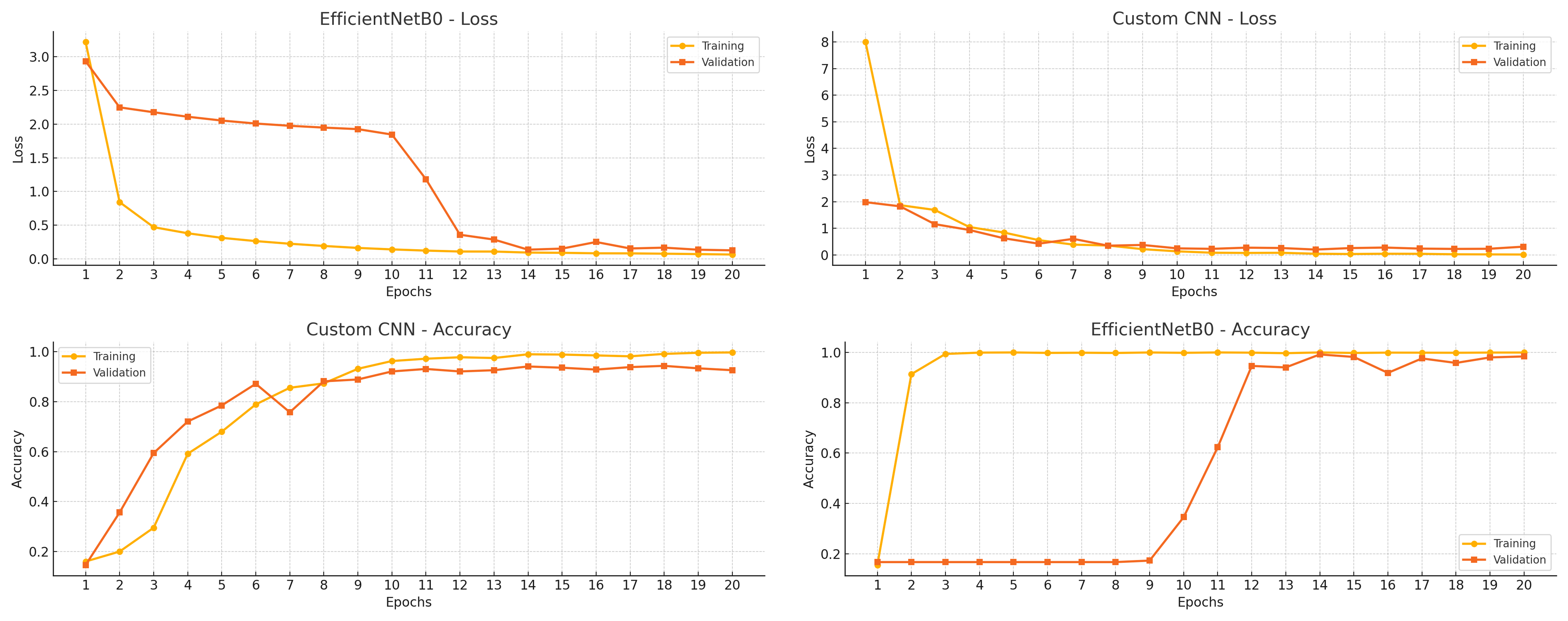}
    \caption{Training and validation accuracy and loss for EfficientNetB0 and Custom CNN models}
    \label{fig:learning_curves}
\end{figure*}

\begin{figure}[h]
    \centering
    \includegraphics[width=0.8\linewidth]{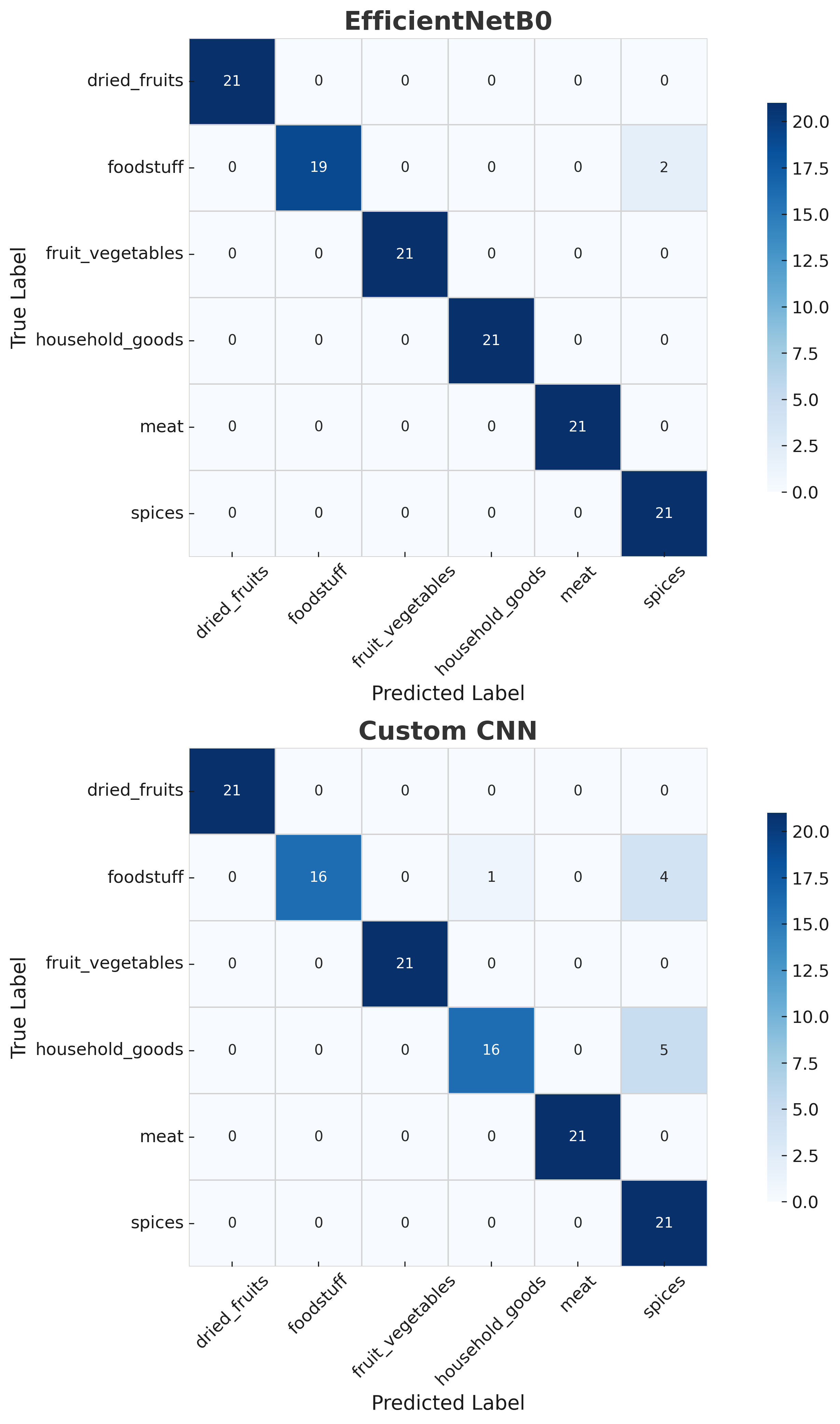}
    \caption{Confusion Matrices for EfficientNetB0 and Custom CNN}
    \label{fig:conf_matrix}
\end{figure}

The results of the tent classification task, comparing the performance of the custom CNN and EfficientNetB0 models. The evaluation is based on accuracy, precision, recall, F1-score, confusion matrix, and mean average precision (mAP). The models were trained and validated on an augmented dataset but tested on the original dataset (126 images) to assess real-world performance.

The classification performance of both models was evaluated using various metrics, including accuracy, loss, precision, recall, and F1-score. Tables~\ref{tab:accuracy_comparison}, \ref{tab:loss_comparison}, and \ref{tab:precision_recall_f1} summarize these values across training, validation, and testing. The mean average precision (mAP) scores for the models showed the Custom CNN achieving 0.98 and EfficientNetB0 achieving 0.99.

\begin{table}[h]
    \centering
    \caption{Comparison of Model Accuracy}
    \label{tab:accuracy_comparison}
    \begin{tabular}{|c|c|c|c|}
        \hline
        \textbf{Model} & \textbf{Train Acc. (\%)} & \textbf{Val. Acc. (\%)} & \textbf{Test Acc. (\%)} \\
        \hline
        Custom CNN & 0.9968\% & 0.9257\% & 0.9206\% \\
        EfficientNetB0 & 0.9984\% & 0.9722\% & 0.9841\% \\
        \hline
    \end{tabular}
\end{table}

\begin{table}[h]
    \centering
    \caption{Comparison of Model Loss}
    \label{tab:loss_comparison}
    \begin{tabular}{|c|c|c|}
        \hline
        \textbf{Model} & \textbf{Validation Loss} & \textbf{Test Loss} \\
        \hline
        Custom CNN & 0.3115 &  0.3635 \\
        EfficientNetB0 & 0.1189 & 0.1082 \\
        \hline
    \end{tabular}
\end{table}

\begin{table}[h]
    \centering
    \caption{Precision, Recall, and F1-score Comparison}
    \label{tab:precision_recall_f1}
    \begin{tabular}{|c|c|c|c|}
        \hline
        \textbf{Model} & \textbf{Precision (\%)} & \textbf{Recall (\%)} & \textbf{F1-score (\%)} \\
        \hline
        Custom CNN & 0.9340\% & 0.9289\% & 0.9296\% \\
        EfficientNetB0 & 0.9855\% & 0.9841\% & 0.9841\% \\
        \hline
    \end{tabular}
\end{table}

EfficientNetB0 consistently outperformed the custom CNN across all datasets, demonstrating higher accuracy and lower loss values.

\subsection{Learning Curves}
To visualize training dynamics, we plotted loss and accuracy curves over epochs. Figure~\ref{fig:learning_curves} presents the learning curves for both models. The learning curves indicate that EfficientNetB0 achieves a sharp improvement in accuracy after the initial epochs, reaching nearly 100\% while maintaining low validation loss, suggesting strong generalization. The custom CNN shows a more gradual increase in accuracy but maintains a noticeable gap between training and validation performance.

\subsection{Confusion Matrix Analysis}
A confusion matrix provides a deeper understanding of model misclassification. Figure~\ref{fig:conf_matrix} illustrates the confusion matrices. The custom CNN frequently misclassified household goods and foodstuff, leading to lower recall. In contrast, EfficientNetB0 reduced misclassification errors significantly, showing better separation between categories.

EfficientNetB0 achieved a higher mAP score, demonstrating better confidence in predictions.
EfficientNetB0 significantly outperformed the custom CNN in accuracy, recall, precision, and F1-score. 

These results validate the effectiveness of transfer learning in tent classification, proving that pre-trained models generalize better than models trained from initialization.

\section{Conclusion}

This study advances bazaar-specific image classification by demonstrating the superiority of transfer learning over custom architectures, offering a scalable tool for market automation. We explored deep learning-based classification of tents in street bazaars, comparing a custom CNN trained from initialization with EfficientNetB0, a pre-trained model utilizing transfer learning. Our results demonstrate that EfficientNetB0 significantly outperforms the custom CNN across multiple evaluation metrics, including accuracy, precision, recall, F1-score, and mean average precision (mAP).

Our findings confirm that transfer learning provides a robust and efficient solution for tent classification, leveraging pre-trained features to achieve high accuracy with limited training data. This demonstrates the practicality of using deep learning for marketplace organization and automation in real-world settings.

While the results demonstrate high performance, this study has several limitations. The dataset includes only six classes of tents, all of which were captured in outdoor environments. This restricts the model’s ability to generalize to a broader range of tent types, such as indoor stalls or more diverse tent designs found in different regional markets. Moreover, the model has not been tested on tents from other bazaars, which may introduce new visual features and environmental conditions. Therefore, the current findings are limited in scope and should be validated further with larger and more diverse datasets.

Future work could involve expanding the dataset to include more tent types, integrating object detection techniques to locate tents within images, and deploying the model for real-time use in mobile or web-based systems for on-site classification and analytics. A potential direction for the future development of the proposed model is its integration into drones and robots, which could automate data collection and labeling by navigating through bazaars or markets \cite{Isaev2024}. In addition, exploring innovative neural network architectures, such as SpinalNet, which has achieved high accuracy in diverse applications \cite{dim2021}, could further improve the efficiency and performance of tent classification models in complex, real-world environments.

\end{document}